\pdfoutput=1

\documentclass[11pt]{article}

\usepackage[preprint]{coling}

\usepackage{times}
\usepackage{latexsym}

\usepackage[T1]{fontenc}

\usepackage[utf8]{inputenc}

\usepackage{microtype}

\usepackage{inconsolata}

\usepackage{graphicx}
\usepackage{adjustbox}

\usepackage{expex}
\usepackage{booktabs}
\usepackage{comment}

\usepackage{graphicx}
\usepackage{subcaption}

\usepackage{xcolor}

\usepackage{forest}

\usepackage{hyperref}

\title{Automatic Extraction of Clausal Embedding Based on Large-Scale English Text Data }

\author{
 \textbf{Iona Carslaw\thanks{These authors contributed equally and are ordered alphabetically.}\textsuperscript{1, 2}},
 \textbf{Sivan Milton\footnotemark[1]{}\textsuperscript{1, 2}},
 \textbf{Nicolas Navarre\footnotemark[1]{}\textsuperscript{1, 2}},
\\
 \textbf{Ciyang Qing\textsuperscript{2}},
 \textbf{Wataru Uegaki\textsuperscript{2}}
\\
\\
 \textsuperscript{1}School of Informatics, University of Edinburgh\\
 \textsuperscript{2}School of Philosophy, Psychology \& Language Sciences, University of Edinburgh
\\
 \small{
   {\{ \href{mailto:I.C.A.Carslaw@sms.ed.ac.uk}{I.C.A.Carslaw}, \href{mailto:s.milton@sms.ed.ac.uk}{s.milton}, \href{mailto:n.s.navarre@sms.ed.ac.uk}{n.s.navarre}\}@sms.ed.ac.uk}\quad {\{\href{mailto:cqing@ed.ac.uk}{cqing}, \href{mailto:w.uegaki@ed.ac.uk}{w.uegaki}\}@ed.ac.uk}
 }
}

\begin{document}
\maketitle
\begin{abstract}
For linguists, embedded clauses have been of special interest because of their intricate distribution of syntactic and semantic features. Yet, current research relies on schematically created language examples to investigate these constructions, missing out on statistical information and naturally-occurring examples that can be gained from large language corpora. Thus, we present a methodological approach for detecting and annotating naturally-occurring examples of English embedded clauses in large-scale text data using constituency parsing and a set of parsing heuristics. Our tool has been evaluated on our dataset Golden Embedded Clause Set (GECS), which includes hand-annotated examples of naturally-occurring English embedded clause sentences. Finally, we present a large-scale dataset of naturally-occurring English embedded clauses which we have extracted from the open-source corpus \textit{Dolma} using our extraction tool.
\end{abstract}

\section{Introduction}
\label{sec:ec}

One of the most popular methods of conducting linguistic research has consisted of handcrafting paradigmatic utterances followed by gathering native speakers' judgements. Yet, it is questionable how much these constructed utterances reflect real-world language use. 
As a result, plenty of debate has arisen about the legitimacy of paradigmatic utterances as a research tool, with arguments suggesting this particular data collection technique can lead to biased results \cite{Cowart, carson}. Whilst this debate has been happening in linguistics, the advancements of Natural Language Processing (NLP) have led to a significant increase in the amount of freely available language corpora as well as an increase in their size. For example, the open-source dataset \textit{Dolma} consists of 3 trillion English tokens \cite{dolma}. 
These datasets provide new opportunities for linguistic research, with the ability to gather statistical data about specific language constructions and 
naturally-occurring examples beyond handcrafted sentences.  

One particular sentence construction that would benefit from such corpus research is that of \textsc{embedded clauses}. These constructions contain an embedding predicate which selects a clausal complement, as seen in the sentence: \textit{Mary hopes that John likes chocolate}. Here, the predicate \textit{hopes} embeds the declarative clausal complement \textit{that John likes chocolate}. Alongside \textsc{declarative} clausal complements, as in (\ref{decl}), there are also \textsc{polar interrogative} clausal complements (\ref{polq}), \textsc{alternative interrogative} clausal complements (\ref{altq}), and \textsc{constituent interrogative} clausal complements (\ref{constq}). Crucially, predicates vary with respect to which clausal complement type they are allowed to embed; consider the difference in grammaticality between \textit{wonder}, which can embed interrogative clausal complements, and \textit{hope}, which cannot embed interrogative clausal complements.\footnote{
These judgements, commonly
reported in the literature, are 
shared by the 3 native British English and Canadian English speakers among the authors.} In addition, it has been observed that emotive factives, such as \textit{be happy (about)}, take declarative and constituent interrogative complements but not polar and alternative interrogative complements \cite[][a.o.]{abels2004surprise, karttunen1977syntax, saebo2007whether}.

\pex 
\a Mary \{*wondered | hoped | was happy \} [that John liked chocolate].\label{decl}
\a Mary \{wondered | *hoped | *was happy about \} [whether John liked chocolate]. \label{polq}
\a Mary \{wondered | *hoped | *was happy about \} [whether John liked chocolate or cake]. \label{altq}
\a Mary \{wondered | *hoped | was happy about \} [which chocolate John ate]. \label{constq}
\xe

Because of this observation that a predicate selects for particular types of embedded clause in fine-grained ways - partly conditioned by the predicate's lexical semantics - there is a debate amongst syntacticians and semanticists about what roles syntax and semantics play within these constructions \cite[][a.o.]{grimshaw1979complement, uegaki2019hope, WhiteAaronSteven2021Obah}. Extrapolating clausal embeddings from large-scale corpora would help to answer such questions, by providing large-scale statistical evidence for how often these embedding predicates appear in natural language use and what clausal complements they select, as well as the ability to look for natural language examples. Thus, the aim of this paper is to create a tool for linguists to extract English sentences containing embedded clauses from large-scale corpora, whilst also providing the following information: (i) the span of the embedded clause, (ii) the lexeme(s) of the embedding predicate, and (iii) the type of the embedded clause.\footnote{Code: \href{https://github.com/navarrenicolas/clause_parser/}{https://github.com/navarrenicolas/clause\_parser/}.\\ 
Extracted embedded clause dataset available on HuggingFace: \href{https://huggingface.co/datasets/nnavarre/Embedded_Clauses-dolma_v1_6-sample}{https://huggingface.co/datasets/nnavarre/Embedded\_Clauses-dolma\_v1\_6-sample}} 

This task of extracting embedded clauses is by no means trivial. Firstly, the span of the embedded clause in a sentence has to be correctly identified, excluding any element that belongs to the matrix clause. Secondly, there are constructions that superficially resemble embedded clauses, but are in fact not, as they fail to categorise syntactically as
\emph{complements} of an embedding predicate or as \emph{clauses}. To see this, consider 
the following examples:
\pex\label{adverse}
\a
Mary saw a man [that John mentioned]. \label{rel}
\a 
Mary ate [what John cooked]. \label{free}
\a 
Mary goes to the gym regardless of [whether she is tired or not]. \label{uncond}
\xe
The bracketed clause in (\ref{rel}) is a \textsc{relative clause}
and is not a complement of an embedding predicate. In (\ref{free}), we have an instance of a \textsc{free relative}, which is considered as primarily a Noun Phrase rather than a clause \cite{caponigro2003free,van2006free}. The bracketed clause in (\ref{uncond}) is an \textsc{unconditional} \cite{rawlins}, which is a modifier rather than a complement of a matrix predicate. Thirdly, embedded clauses can arise in complex clausal structures such as coordination (\ref{coor}), which often occurs with ellipsis, nesting (\ref{nest}), or some combination of both (\ref{coor_nest}). Consequently, to correctly identify embedded clauses, we need a correct syntactic parse of the sentence, as well as appropriate heuristics to rule out structures such as those in (\ref{adverse}) and deal with the structures in (\ref{complex}).

\pex\label{complex}
\a
Mary knows [that John likes chocolate] and [that Mark does not]. \label{coor}
\a 
Mary knows [that John thinks [that Mark likes chocolate]]. \label{nest}
\a 
Mary knows [that John thinks [that Mark likes chocolate]] and [that Mark does not]. \label{coor_nest}
\xe



Our paper is structured in the following way: Section \ref{sec:rw} describes previous attempts at building a large-scale corpora of English embedded clauses (e.g.\ MegaAcceptability), and additionally examines existing tools designed to extract sentences from language corpora (e.g.\ linguistic search engines). Section \ref{sec:goldenset} introduces our hand-annotated dataset of English embedded clauses: Golden Embedded Clause Set (GECS). Section \ref{sec:m} describes our extraction tool that uses constituency representations and parsing heuristics, as well as our tool's performance on GECS. Section \ref{sc:large-scale-dataset} presents the large English embedded clause dataset that we have extracted from the open-source dataset \textit{Dolma}. Section \ref{sec:fr} suggests future research avenues and \ref{sec:conclusion} concludes our work. Overall, we provide three new contributions: 

\begin{enumerate}
    \item A small-scale dataset (GECS) with fine-grained gold standard annotation of English embedded clauses to be used as a benchmark for this task
    \item An extraction tool which can be applied to English language corpora to extract and annotate embedded clauses
    \item A large-scale extracted set of English embedded clauses from the language corpus \textit{Dolma} for the linguistic community to use
\end{enumerate}

\section{Relevant Work}
\label{sec:rw}

\subsection{MegaAcceptability}
\label{sec:MA}
The only existing attempt at a large dataset of English embedded clauses is the MegaAcceptability dataset \cite{WhiteAaronSteven2016Acmo, WhiteAaronSteven2020Faas}. White and Rawlins selected a list of $1007$ English verbs that are known to select clausal embeddings, and then designed $50$ schematic sentences covering a range of syntactic environments in which an embedded clause can occur. They then slotted the $1007$ verbs into the $50$ schematic sentences to create $\sim50,000$ entries. Through Amazon MTurk, participants rated the acceptability of the resultant sentences, leading to a large dataset of embedded clause constructions ranked by acceptability, on a $7$-point ordinal scale.

Although the MegaAcceptability dataset moves away from the problem of a small set of sentences being used as evidence for linguistic hypotheses, it still utilises non-natural sentences which have been handcrafted. Furthermore, for finite embedded clauses White and Rawlins (\citeyear{WhiteAaronSteven2016Acmo, WhiteAaronSteven2020Faas}) only considered environments without complementisers or with the following complementisers: \textit{that, whether}, and \textit{which}. They also only consider predicates with no prepositions or with the following prepositions: \textit{to} and \textit{about}. They make use of a pre-defined list of verbs which accept clausal complements, which does not account for the full set of embedding verbs nor adjectives and complex predicates which can also accept clausal complements. Therefore, it is unclear if the dataset captures the natural distributions of embedding predicates, embedded clause types, and the types of embedded clauses selected by embedding predicates.  

\subsection{Linguistic Search Engines}
The goal to extract sentences with certain linguistic phenomena from natural language use is not a new concept. There have been several attempts to create \textit{search engines} in which an individual can query annotated natural-language corpora for certain constructions and then be provided with a list of sentences which match the provided query. Prominent tools with this use include the Linguist's Search Engine \cite{lse}, SPIKE \cite{ShlainMicah2020Ssbe}, and the LINDAT/CLARIAH-CZ PML Tree Query \cite{11858/00-097C-0000-0022-C7F6-3}.

Although these are powerful tools, their query languages are not sufficiently fine-grained to capture the relevant structures of embedded clauses. They rely on annotation of corpora with lemmas, part-of-speech tags, and dependency graph representations. This means that one would need to specify dependency relationships rather than constituency/hierarchical ones to identify the structure of embedded clauses. Such an approach is limiting, as it is difficult to identify clause and predicate spans based on dependency relations or linear structure. There is also less consistency with respect to the relations that identify embedded clauses than with constituency parsers. Moreover, linguistic search engines offer linguists limited flexibility to decide which corpora they want to extract sentences from.

\section{Golden Embedded Clause Set (GECS)}
\label{sec:goldenset}

\ifx
\begin{figure*}
\begin{subfigure}[t]{0.5\textwidth}
    \centering
    \includegraphics[width=\linewidth]{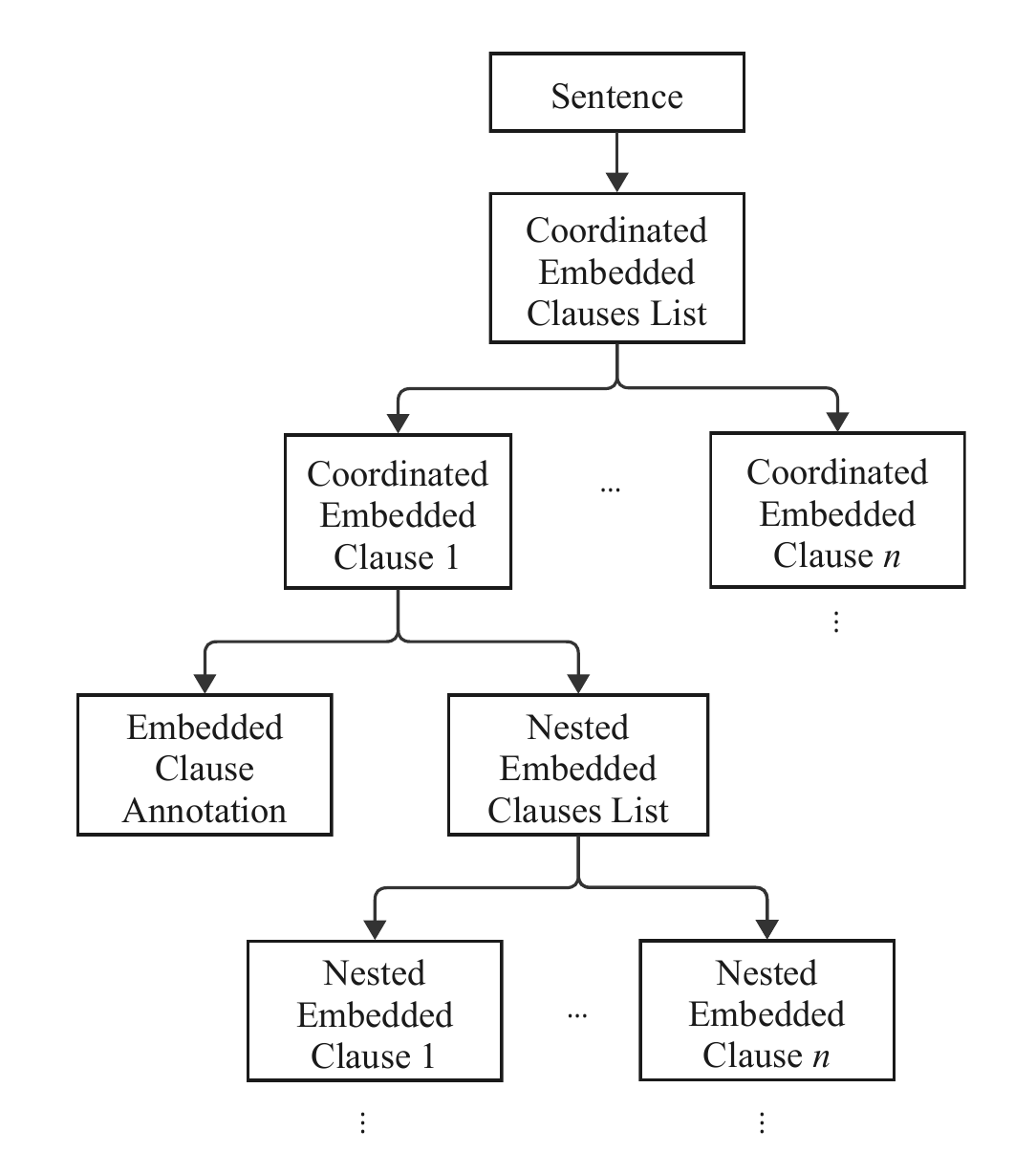}{}
      \caption{The annotation in GECS for each embedded clause.}
      \label{fig:struct}
\end{subfigure}
\begin{subfigure}[t]{0.5\textwidth}
    \includegraphics[width=\linewidth]{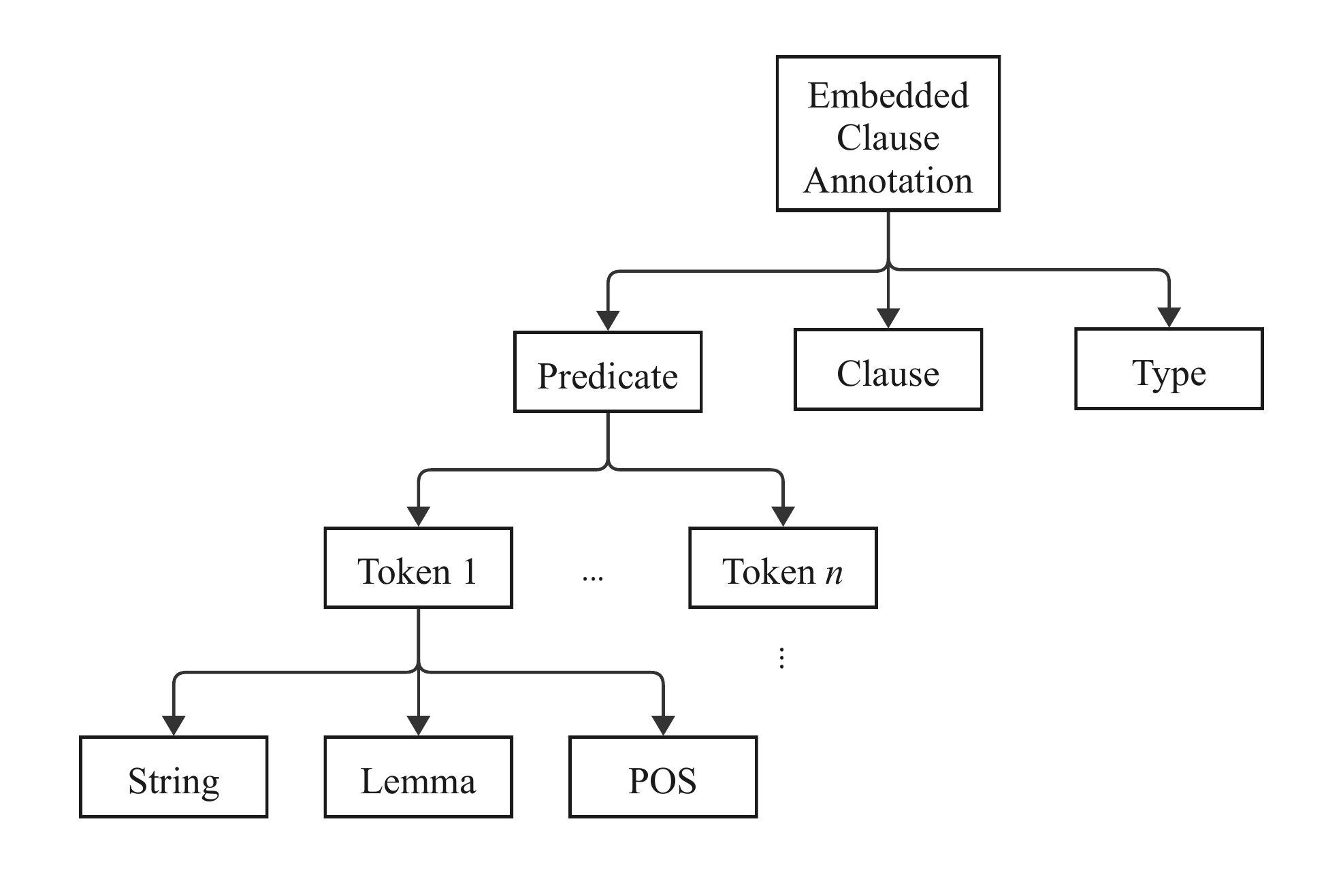}{}
      \caption{The data structure in GECS for coordination and nesting of embedded clauses.}
      \label{fig:annot}
\end{subfigure}
\end{figure*}
\fi

For the novel task of English embedded clause detection in natural language corpora, we created a hand-annotated dataset (GECS) which can serve as a benchmark for evaluation and be used in its own right for a small-scale analysis of embedded clause constructions. 
In GECS, each embedded clause is annotated with its embedding predicate, clause span, and clause type (see Figure \ref{fig:annot}). We provide the embedding predicate as a list of the relevant tokens (i.e.\ ignoring negation words, adjuncts, and any other tokens which may appear between the first embedding predicate token and the clause).\footnote{To capture the complex constructions of coordinated and nested embedded clauses alluded to in Section \ref{sec:ec}, we optionally provide a recursive data structure version of GECS which makes the internal clausal structure transparent (see Figure \ref{fig:struct} in the Appendix). }

\begin{figure}
\centering
  \includegraphics[width=8cm]{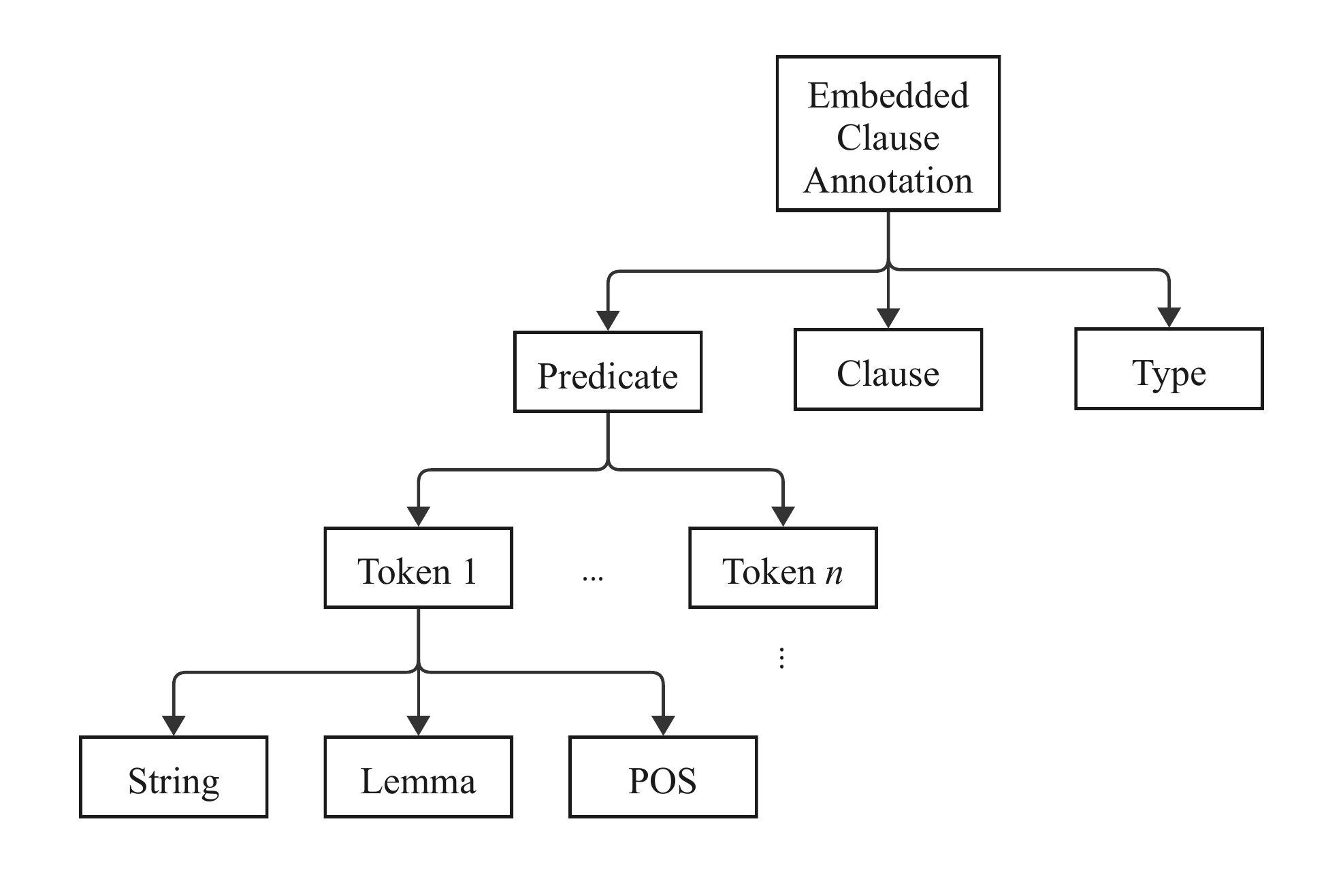}{}
  \caption{The annotation in GECS for each embedded clause.}
  \label{fig:annot}
\end{figure}

\begin{figure*}
    \begin{subfigure}{0.45\linewidth}
        \begin{adjustbox}{width=\linewidth}\begin{forest} [S [NP [PRP We]] [VP [VBP are] [ADVP [RB still]] [ADJP [JJ unclear] [PP [IN as] [PP [IN to] [SBAR [IN whether] [CC or] [RB not] [S [NP [NNP Canada]] [VP [VBZ is] [NP [DT a] [NN state]]]]]]]]]] \end{forest}\end{adjustbox}
        \caption{Complex embedding predicate (`are unclear as to')}
        \label{fig:parse1}
    \end{subfigure}
    \hfill
    \begin{subfigure}{0.5\linewidth}
        \begin{adjustbox}{width=\linewidth}\begin{forest} [S [NP [DT No] [PRP one]] [VP [VBZ is] [ADJP [JJ certain]] [SBAR [IN whether] [S [NP [NNP Betty]] [VP [VBZ is] [ADVP [RB still]] [ADJP [JJ alive]] [CC or] [VP [ADVP [RB just]] [VBN hypnotized]]]]]]] \end{forest}\end{adjustbox}
        \caption{Adjectival predicate (`is certain')}
        \label{fig:parse2}
    \end{subfigure}
    \caption{Example parses of embedded clause sentences in GECS.}
    \label{fig:parses}
\end{figure*}
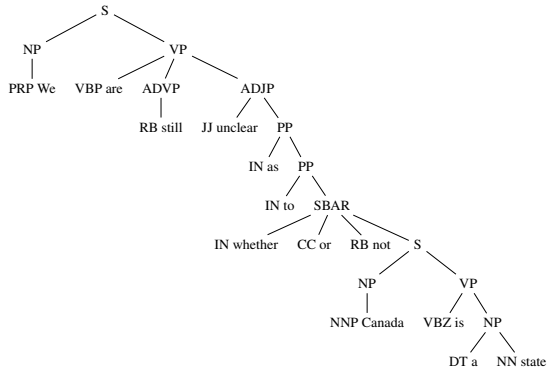
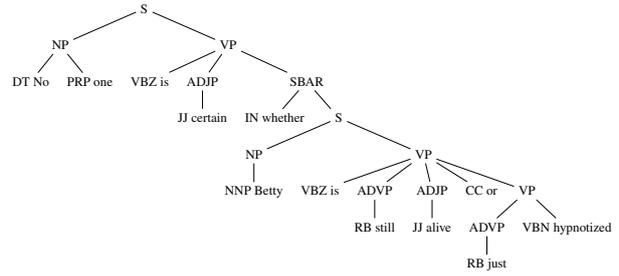

\paragraph{Annotation Procedure}
To create our naturally-occurring embedded clause dataset, we selected a subset of $866,538$ sentences from \textit{Dolma}\footnote{Specifically we used the files \textit{cc\_en\_head-0000}, \textit{cc\_en\_head-0001}, and \textit{c4-0085}.} \cite{dolma}. The data was not cleaned so as to accurately test the robustness of the tool. We then parsed the sentences and filtered them to remove any which necessarily did not contain embedded clauses.\footnote{We used SBAR from SpaCy's Berkeley Neural Parser \cite{kitaev-etal-2019-multilingual,kitaev-klein-2018-constituency} as an indicator, assuming that this structure in the parsed representation is a necessary condition for an embedded clause.} To extract the set of polar and alternative interrogative embedded clauses, we further filtered out sentences that did not contain the words \textit{whether} or \textit{if}. Finally, to filter for constituent interrogative embedded clauses, we only considered sentences with: \textit{who}, \textit{what}, \textit{when}, \textit{where}, \textit{why}, \textit{how}, or \textit{which}. The next stage of hand annotation consisted of one researcher going through the pre-filtered sentences and confirming (i) if there were embedded clauses and (ii) if so, providing the annotation of predicate tokens, clause span, and type. A second researcher then went through the previous researcher's annotations to confirm agreement. 

Overall, GECS contains $147$ declarative embedded clauses, $138$ polar interrogative embedded clauses, $84$ alternative interrogative embedded clauses,\footnote{There are fewer alternative interrogative embedded clauses due to this type being far sparser in the pre-filtered dataset than the other types.} and $158$ constituent interrogative embedded clauses. In addition, we provide a set of $111$ adversarial examples verified to not contain any embedded clauses, but do contain misleading structures such as free relatives and relative clauses. These were created by selecting sentences discarded by the annotators in the final stage of GECS' creation.

\section{Parser Tool}
\label{sec:m}

Though it is possible to define a set of heuristics based on Regular Expressions or dependency relations, preliminary analysis indicates significant disadvantages to such an approach, as were seen with the linguistic search engines from Section \ref{sec:rw}. For this reason, we opted for representations from constituency syntactic parsers to extrapolate hierarchical structure. A benefit of this choice is that linguistic theory is typically given with respect to constituency trees, and we can therefore implement linguistic facts into extraction heuristics more freely than with other representations. While it is possible that a Dependency Parser could be used to achieve equivalent results, it is not clear what improvements it could offer. We leave this question open to a more thorough exploration in future. With the constituency representation we defined a set of heuristics to perform the following tasks: 
\begin{enumerate}
\setlength{\itemsep}{0pt}
\setlength{\parskip}{0pt}
\setlength{\parsep}{0pt}
    \item \textbf{Detection}: detecting embedded clause(s) in a sentence
    \item \textbf{Predicate Identification}: identifying each embedding predicate
    \item \textbf{Clause Identification}: identifying the span of each embedded clause
    \item \textbf{Typing}: identifying the type of each embedded clause
\end{enumerate}

The syntactic parser that we use is SpaCy's Berkeley Neural Parser, a constituency parser that has an LSTM and self-attentive architecture \cite{kitaev-klein-2018-constituency, kitaev-etal-2019-multilingual}. Other options are available for constituency parsing; however, we decided upon this parser because it is state-of-the-art for constituency parsing.

The SpaCy constituency parser represents each sentence as an n-ary tree structure with several syntactic categories (e.g. S, VP, NP, SBAR) in parent and child hierarchy. This tree structure is particularly helpful in extracting embedded clauses because we can traverse the parent levels and check for particular child nodes in complement positions. 
We then defined heuristics based on the structures from the parser to perform the aforementioned tasks of embedded clauses detection, predicate identification, clause identification, and typing. 

\subsection{Methodology}

\paragraph{Detection}
The first heuristic we deemed necessary for detecting embedded clauses is the existence of an SBAR in the parsed representation. This is a syntactic category for a subordinate clause, which is a superset of embedded clauses, but also includes non-embedded clauses like relative clauses. To check if a subordinate clause is an embedded clause, we assume it needs to be dominated by a VP headed by a predicate. While there may be other syntactic categories immediately above the subordinate clause, we are only interested in the first upstream occurrence of one of two syntactic categories: NP or VP. In the case where the label is VP, the sentence has an embedded clause. If the label is NP, then the sentence does not have an embedded clause---likewise if neither of the two are found until the root node of the tree. We use the hierarchical nature of the constituency parser to distinguish embedded clauses from relative clauses and complements of NPs.

To limit the amount of false positives that would be extracted from the dataset we implemented a few heuristics based on the embedding predicate and the subordinating conjunction of the clauses that are detected.
First, if the embedding predicate is empty after the part-of-speech filtering or the only predicate token is `is/be', then the clause is not considered to be an embedded clause.
Secondly, we rule out any clauses beginning with certain subordinating conjunction because they are not indicative of an embedded clause. Specifically, we blacklist the following: \textit{after, although, before, despite, to, for, so, though, unless, until, than, because, since, while, as, even if, in order}.

\paragraph{Predicate Identification}
Having identified an embedded clause in a sentence, we can extract the embedding predicate from the sentence by searching for the nearest VP parent of the clause. We iteratively search through the parents of the embedded clause until a VP parent is reached. We then identify the predicate span from this constituent, considering a wider range of possible verbs, adjectives, and prepositions than previous methods (cf. Section \ref{sec:MA}).
For each constituent child of the VP (with exception to the final one which contains the embedded clause) we keep every token in the child span as long as the child label is either a PP, NP or SBAR label.
For the last child of the VP, we keep every token until the onset of the embedded clause.
We then filter these tokens based on their part-of-speech tags.
We keep only the tokens that are VERB, ADP or ADJ, with the exception for an auxiliary tag AUX if there is also an adjective in the original token list.
This helps us capture adjectival predicates such as `unclear as to' or `is certain' (see Figure~\ref{fig:parses}).

\paragraph{Clause Identification}
Given that a sentence is detected as having an embedded clause, we can then further use the parsed representation to extract the span of the embedded clause. The constituency parser is advantageous in this regard as we take whatever is under the syntactic label of SBAR to be the embedded clause constituent. 

\paragraph{Typing}
Having identified the clause span, the heuristics for typing the clause can involve more simple string matching. For alternative interrogative clauses we check the complementiser. If the complementiser \textit{whether} is in the first word of the embedded clause along with the token \textit{or},  then it is an alternative interrogative. If instead, we find \textit{whether} that is not followed by the token \textit{or} or is followed by the explicit string \textit{or not}, then the embedded clause is a polar interrogative. If a unique token of either \textit{which, who, what, when, where, why,} or \textit{how} is the first word of the embedded clause, then it is a constituent interrogative clause. If none of the prior conditions are met, including if the clause begins with \textit{that}, then we type the clause as declarative.

\subsection{Evaluation}
\label{sec:eval}

We evaluate the performance of our tool on the sentence annotations in GECS. With these annotations we can accurately test the tool's ability to detect embedded clauses, embedding predicates, and clause types, allowing us to evaluate how our tool handles messy natural data. We have also built a pattern matching baseline to compare our heuristics against a more linear approach.

\paragraph{Pattern Matching Baseline}

The baseline we constructed is a rule-based tool using pre-defined lexical patterns to extract embedded clause annotations from a sentence. This method relies on the SpaCy Matcher, a tool which is similar to Regular Expressions in that it matches a given pattern in a string, but with useful supplemental linguistic information encoded, such as POS and lemma \cite{spacy2}\footnote{We opted for a Regular Expression matcher baseline due to its straightforward implementation for this task, as compared with a Dependency Parser for instance.}. In order to detect embedded clauses, the Matcher is provided with the fixed list of (potentially) embedding predicates from MegaAcceptability \cite{WhiteAaronSteven2016Acmo}. It then returns instances of these predicates in a sentence; with an added heuristic ensuring that the predicate is followed by some other verb or auxiliary (i.e. a clause), an embedded clause is identified. Prepositions proceeding the verb are included in the list of predicate tokens and POS and lemmas are also identified. Limited by the linear nature of the Matcher, we define the clause span as the end of the predicate to the end of the sentence, ignoring any adverb/pronoun which may occur between a predicate and clause. For the final goal of typing the embedded clause we again use the Matcher to match the first token of the clause to the associated complementisers for each type. We distinguish between polar and alternative interrogatives by classifying clauses containing the token \textit{or} but not the string \textit{or not} as alternative, and every other instance as polar. If no associated complementiser is found in the clause, the clause is typed as declarative.

\paragraph{Results}
\ifx
\begin{table*}
\resizebox{\textwidth}{!}{
\begin{tabular}{@{}lcccccc@{}}
\toprule
                     & \textbf{\begin{tabular}[c]{@{}l@{}}Detection \\ (Single)\end{tabular}} & \textbf{\begin{tabular}[c]{@{}l@{}}Detection\\ (Multi)\end{tabular}} & \textbf{\begin{tabular}[c]{@{}l@{}}Detection \\ (Overall)\end{tabular}} & \textbf{\begin{tabular}[c]{@{}c@{}}Predicate \\ Identification \end{tabular}} & \textbf{\begin{tabular}[c]{@{}c@{}} Clause \\ Identification \end{tabular}} & \textbf{\begin{tabular}[c]{@{}c@{}}Type \\ Identification \end{tabular}} \\ \midrule
\textbf{Baseline}    & 0.52                                                                   & 0.65                                                                 & 0.53                                                                    & 0.79               & 0.50            & 0.94          \\
\textbf{Parser Tool} &0.85&0.79& 0.85 &      0.91         &          0.87       &        0.96      \\ \bottomrule
\end{tabular}}
\caption{Accuracy scores for clause detection in the Baseline and Parser Tool. The identification tasks are evaluated relative to the true positive clause detections.} 
\label{tab:accuracy}
\end{table*}
\fi

\begin{table}[h]
\begin{adjustbox}{width=1\columnwidth}
\begin{tabular}{@{}lllllll@{}}
\toprule
\textbf{Detection}              & \multicolumn{3}{c}{\textbf{Baseline}} & \multicolumn{3}{c}{\textbf{Parser Tool}} \\ \midrule
                       & Precision     & Recall     & F1       & Precision       & Recall       & F1      \\
\textbf{Single} & 0.54          & 0.94       & \textbf{0.69}     &     0.90            &  0.94            &  \textbf{0.92}       \\
\textbf{Multi}  & 0.74          & 0.85       & \textbf{0.79}     &       0.94          &   0.83           &   \textbf{0.88}      \\
\textbf{Overall}       & 0.54          & 0.91       & \textbf{0.68}     &      0.90           &   0.91           &    \textbf{0.91}     \\ \bottomrule
\end{tabular}%
\end{adjustbox}
\caption{Precision, Recall, and F1 scores for clause detection in GECS.} 
\label{tab:detection}
\end{table}

\begin{table}[h!]
\resizebox{\columnwidth}{!}{%
\begin{tabular}{lcc}
\hline
\textbf{Task}                     & \textbf{Baseline} & \textbf{Parser Tool} \\ \hline
\textbf{Predicate Identification} & 0.79              & 0.91                 \\
\textbf{Clause Identification}    & 0.50              & 0.87                 \\
\textbf{Type Identification}      & 0.94              & 0.96                 \\ \hline
\end{tabular}%
}
\caption{Identification accuracy scores evaluated on the true positives set of detected clauses from Table \ref{tab:detection}.} 
 \label{tab:task_results}
\end{table}



We split the evaluation of our tool and the baseline into three detection sections: \textit{Single Clause Evaluation} which evaluates detection performance on sentences in GECS that only included one embedded clause, \textit{Multi Clause Evaluation} which evaluates detection performance on sentences in GECS that had multiple embedded clauses (nested and coordinated clauses), and \textit{Overall} which combines the statistics of single and multi clause evaluation and performance on the adversarials. Table~\ref{tab:detection} provides the precision and recall for these metrics.  We also evaluated amongst the correctly detected embedded clauses the annotation abilities of our tool and the baseline, by seeing if the selected predicate is correct (\textit{Predicate Identification}), if the selected clause is correct (\textit{Clause Identification}), and if the typing of the clause is correct (\textit{Type Identification}). Table~\ref{tab:task_results} provides the accuracy scores for these metrics. 

As Table~\ref{tab:detection} and ~\ref{tab:task_results} shows, we outperform the baseline in every metric, indicating that our method of utilising a constituency based tool is better than a linear based approach. Our tool only slightly degrades in detection recall when given a sentence that had nested and/or coordinated embedded clauses.

\paragraph{Failure analysis}

In the few cases of our tool's error, we see the following categories: parser errors, unconditionals mistaken as embedded clauses, and incomplete complex predicate detection. The parser error was the biggest issue for failed cases - unfortunately this is an unavoidable error given that any parser will be imperfect. Unconditionals also proved a problem because they are parsed the same as an embedded clause, and therefore are impossible to differentiate from one another. Finally, complex predicates were sometimes incompletely detected so not all of the predicate tokens are placed in the entry. Given that some of these errors are unavoidable, coupled with the tool's high precision and recall, we still take the results to indicate that our tool can be used to create a large-scale dataset of naturally-occurring embedded clauses, as long as researchers propagate the error into their analysis - something which needs to be done with any corpus study.

\section{Case Study: Large-Scale Dataset}
\label{sc:large-scale-dataset}

Having designed a tool which can identify and annotate embedded clauses, we applied it to an English corpus to create a large-scale dataset of annotated embedded clauses. We chose to apply the tool to a subset of \textit{Dolma}\footnote{We extracted text from the the Dolma subset \textit{v1\_6-sample}.} \cite{dolma}. Overall, $28,968,073$ embedded clauses were detected.

\subsection{Comparison with MegaAcceptability}

\begin{figure}
    \centering
    \includegraphics[width=\linewidth]{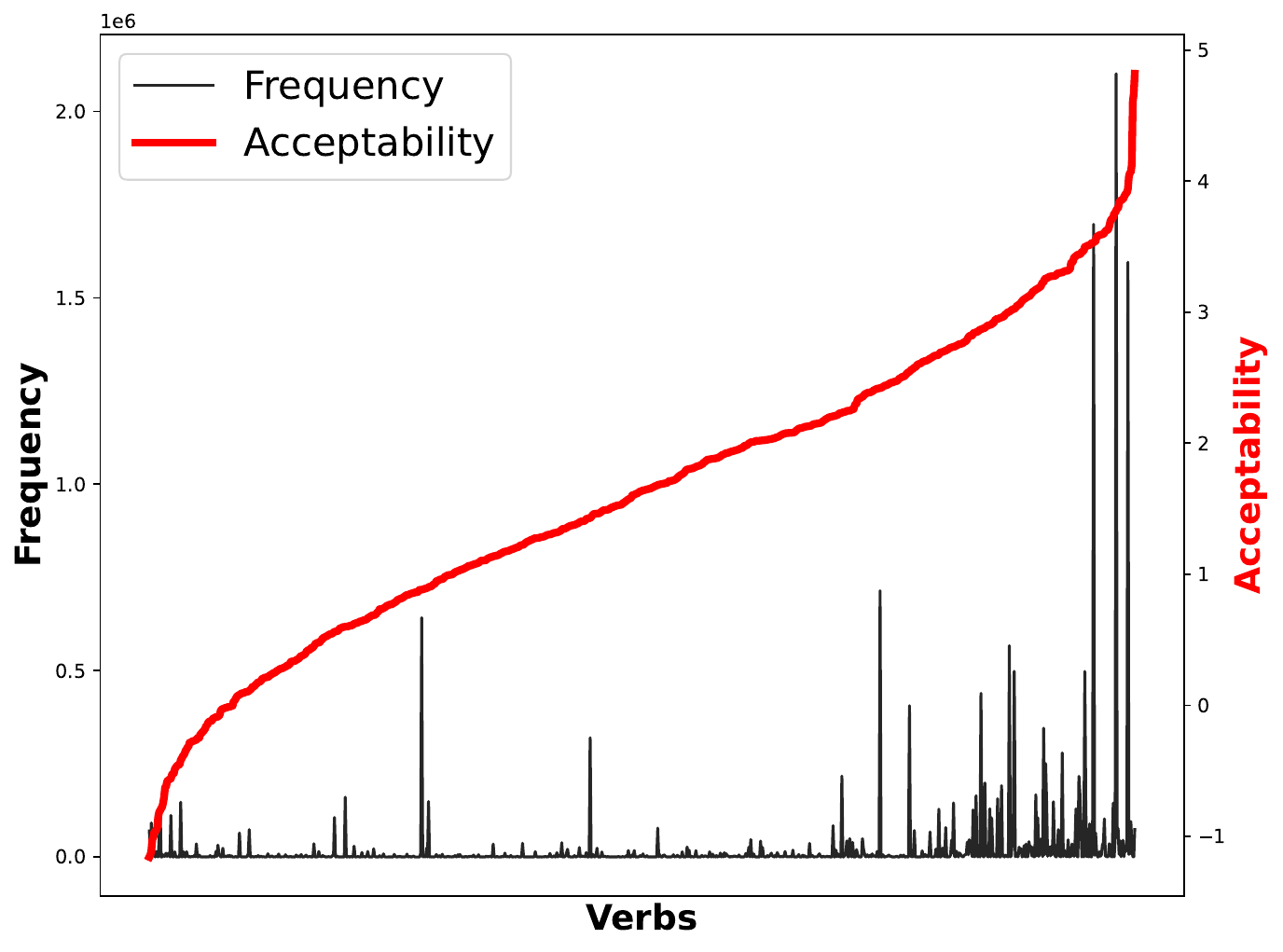}
    \caption{Comparison of natural data frequency and acceptability of the verbs found in MegaAcceptability ranked in increasing order of acceptability}
    \label{fig:freq_acc}
    
\end{figure}

In order to compare with MegaAcceptability, we performed a limited case study on our large-scale dataset by only looking at our dataset entries that included the $1007$ verbs that were used in the MegaAcceptability templates. To get the rating of each verb from MegaAcceptability, we selected the maximum normalised rating of that verb's available constructions. We compared the distribution between the acceptability rating of a verb according to MegaAcceptability and its frequency in the large-scale dataset. It would be generally expected that the higher a verb is rated the more frequent it would be. As shown in Figure \ref{fig:freq_acc}, this is the overall trend that we see. This means that our tool has successfully captured the verbs with the highest acceptability, while the verbs with lower acceptability had a lesser chance of occurring with embedded clauses. 

There are some exceptions to the frequency-acceptability distribution, however this provides an interesting exploration point. For instance, the low acceptability outlier which has a high frequency in Figure \ref{fig:freq_acc} is the predicate \textit{mean}. Looking at entries with \textit{mean} as the predicate, we see three example types: (i) where it is unclear if the predicate is actually embedding or is acting as some filler (\ref{meanfill}), (ii) false positives (\ref{meanfalse}), and (iii) true embedded clauses (\ref{meantrue}). Thus, \textit{mean} could be an outlier because of false positives, or it could be an outlier due to a data-driven approach collecting sentence clause types which a template approach could not.   

\pex
\a It's pretty catchy, I mean who doesn't go ANN ANN and A SORE. \label{meanfill}
\a In Glosa it means "what I've just said". \label{meanfalse}
\a This means [...], the ADA applies to you. \label{meantrue}
\xe

\subsection{Clauses and Predicates at Scale}

With our large scale dataset of embedded clauses we can look beyond the fixed list of predicates as would be provided by a template-driven dataset like MegaAcceptability.
With our approach we are able to view the clause-predicate distribution at a grand scale to test and verify linguistic theories. From the nearly 29 million embedded clause examples in the dataset we have the following distribution of clause types: $19,195,112$ declarative clauses, $9,402,868$ constituent interrogative clauses, $261,274$ polar interrogative clauses, and $108,819$ alternative interrogative clauses. This shows us how rare polar and alternative interrogative clauses are. Moreover, we can examine to the distribution of embedding predicates in the dataset. Taking a look at the part-of-speech tags for each of the embedding predicates in the dataset we can observe the distribution of adjectival and verbal predicates. Adjectival predicates require an accompanying verb or auxiliary (e.g., \textit{be happy}), so we look at complex predicates involving two or more tokens. We find that there are $35,294$ unique adjectives within these complex predicates. Meanwhile, for simple one-word verbal predicates, we find $29,654$ unique predicates. Altogether, this leaves us with a strong set of examples to analyse any clause-predicate distribution of interest.

Here we present an example of how the dataset can be used in linguistic research to further validate and verify linguistic theories, as well as survey new possible sentence constructions that could be of interest.

\paragraph{Emotive Factive Predicates} As mentioned in Section \ref{sec:ec}, previous analyses have shown that emotive factive predicates, such as \textit{be happy}, or \textit{be glad}, are not able to embed either polar or alternative interrogative clauses \cite{karttunen1977syntax,abels2004surprise,saebo2007whether}. We can see if the extracted dataset shows this distribution statistically and if there are any counter-examples. 

To test this generalisation, we selected a subset of emotive factive predicates to investigate further: \textit{happy, amazed, sad, glad, excited, surprised, incredible, angry, mad, jealous, afraid}. Looking at the clauses that are embedded with these predicates, we get the following distribution of clause types: $175,479$ declarative clauses, $47,877$ constituent interrogative clauses, $159$ polar, and $134$ alternative interrogative clauses. The statistical breakdown does match the generalisation, with declarative and constituent interrogative embedded clauses being the more popular embedded clause type. However, more importantly, there are some polar and alternative interrogative, of which we can search through to find potential counter-examples to the generalisation. 

In searching through the polar and alternative interrogative embedded clauses, many are false positives, with the following four errors being indicative of the set: unconditionals (\ref{unconditionalemotive}), wrong predicate span where the emotive factive is not the embedding predicate (\ref{predicateemotive}), real embedded clauses but the sentence does not have
the intended meaning 
required by
the generalisation, e.g., \emph{be afraid} is non-factive in (\ref{realecemotive}), and clausal adjuncts (\ref{adjunctemotive}). 

\pex
\a It’s not your problem, because you’re happy whether you’re with him or doing stuff on your own. \label{unconditionalemotive}
\a I'm not sure how excited to get about this fund and whether he's just piggybacking on the Buffett name. \label{predicateemotive}
\a We are afraid whether it will be in Sindhi interest. \label{realecemotive}
\a Meanwhile, people across the state are hair-on-fire mad over whether urban water users should be allowed to buy rural property simply for the water rights, and whether some water users should be allowed to sell their water to others out of state. \label{adjunctemotive}
\xe 

Given that we need to propagate the tool's error rate, this is to be expected. However, there appears to be some genuine counter-examples (\ref{counter}), of which at least two of the three native speakers among the authors find grammatical. It is beyond the scope of this paper to provide an analysis of these sentences, so we leave it for future work.  

\pex \label{counter}
\a In the post you talk about your child's health issues and in the end ask if people are happy with whether they're circumcised or not. \label{counter1}
\a You might be surprised about whether there’s hope for future shooters.\label{counter2}
\xe

Although this analysis is not exhaustive in the least, we use these examples to motivate the use of this dataset to further validate and explore linguistic theories through naturally-occurring linguistic data in addition to handcrafted templatic examples.

\section{Discussion}
\label{sec:fr}

As this is the first method at extracting embedded clauses from natural language corpora, we set out some future research avenues to be undertaken. Firstly, clausal embedding extraction should be extended to other languages so that linguistic theories using such large corpora can have crosslinguistic validity. Given that the universal definition of a clausal complement is a complement to VP, we argue that a similar method to what we have described in the paper can be taken with other languages.  The main changes would be to the fine-grain heuristics that we used for typing. Of course, our approach is subject to the limitations that follow from any corpus-based research, which introduces its own set of biases pre-existing in the corpora. It is also not always possible to scale this approach crosslinguistically, as the method relies on a given language having large enough corpora (which many do not). Nonetheless, this should not deter people from using the method with an applicable language, in complement with other approaches. Secondly, we recognise that there are other potential methods for extracting English clausal embeddings. One such technique is the use of an LLM, a method which we decided against given that an LLM is a blackbox, meaning a thorough error analysis would not be able to be conducted. To aid future development within this area, we have provided GECS to be used as a benchmark for this task.   

\section{Conclusion}
\label{sec:conclusion}

The availability of large natural-language corpora has led to an opportunity for linguists to conduct large-scale language studies. However, extracting specific language constructions from such large corpora is a difficult endeavour. A particular construction in need of a large-scale corpus study is that of embedded clauses. Thus, we have made three contributions in aid of fulfilling this need. Firstly, we have created GECS, a small-scale dataset with fine-grained gold standard annotation of embedded clauses to be used as a benchmark for embedded clause extraction in English. Secondly, we created a tool which can be applied to English natural language corpora to detect and annotate embedded clauses. And finally, we provided a large-scale extracted set of English embedded clauses from the natural language corpus \textit{Dolma} for the linguistic community to use. 

\section*{Acknowledgements}

This work was supported in part by the UKRI Centre for Doctoral Training in Natural Language Processing, and funded by the UKRI (reference: EP/S022481/1) to the first three authors and the UKRI Future Leaders Fellowship (reference: MR/V023438/1) to the last author. We would like to thank the three anonymous reviewers, Mark Steedman, and Aaron White for their insightful comments.

\bibliography{coling_latex}


\appendix
\section{Data Structure}
\label{sec:appendix}

\begin{figure}[h!]
  \includegraphics[width=7cm]{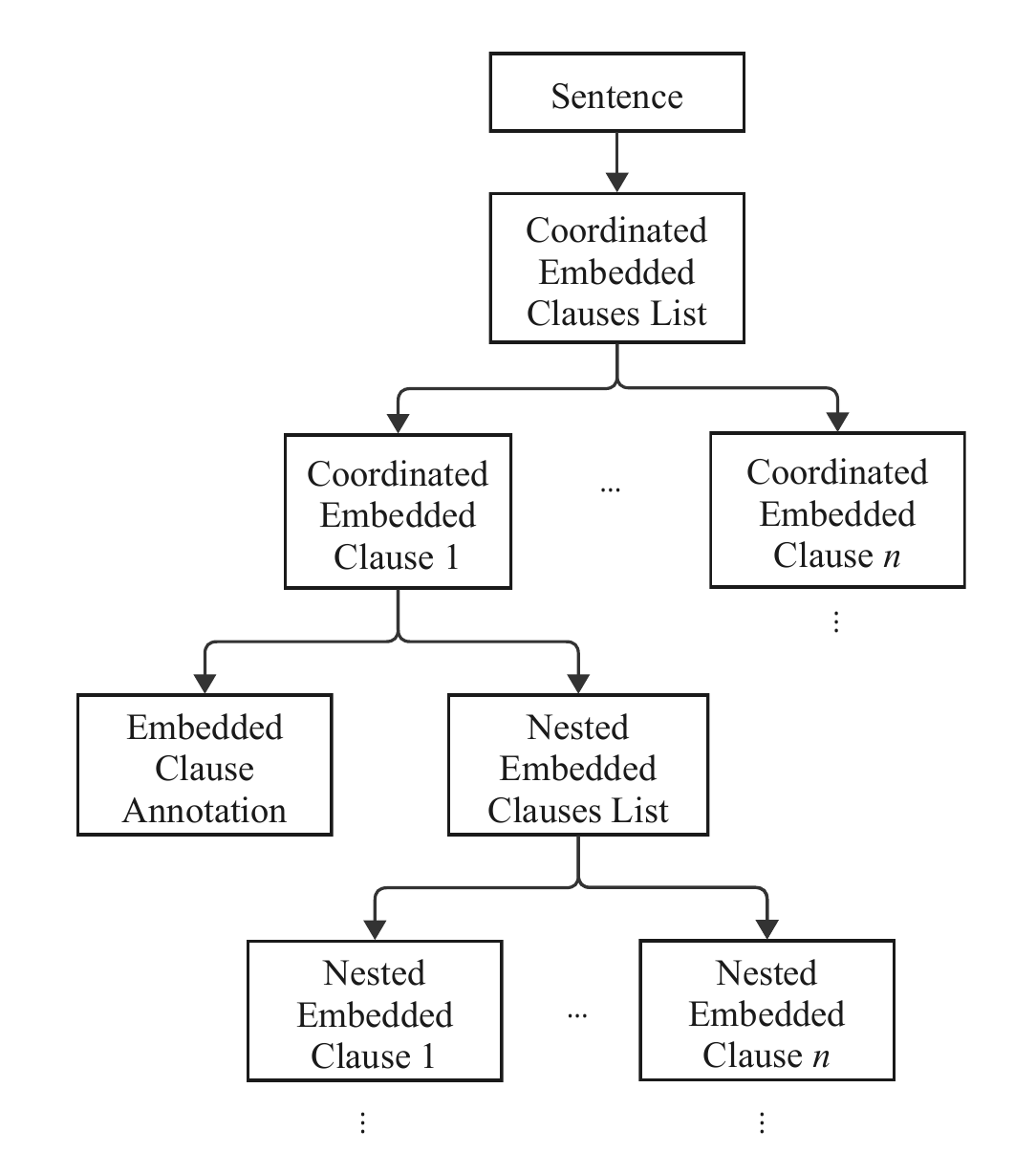}{}
  \caption{The data structure in GECS for coordination and nesting of embedded clauses.}
  \label{fig:struct}
\end{figure}

\section{Examples of Parser Errors}

\begin{figure}
        \begin{adjustbox}{width=\linewidth}\begin{forest} [S [NP [PRP We]] [VP [VBP are] [ADJP [JJ happy] [S [VP [TO to] [VP [VB discuss] [NP [NP [PRP\$ your] [NN case]] [CC and] [SBAR [WHADVP [WRB how]] [S [NP [PRP we]] [VP [MD can] [VP [VB help] [NP [PRP you]]]]]]]]]]]]] \end{forest}\end{adjustbox}
        \caption{False-negative parse of a constituent embedded clause}
        \label{fig:false-neg}
\end{figure}
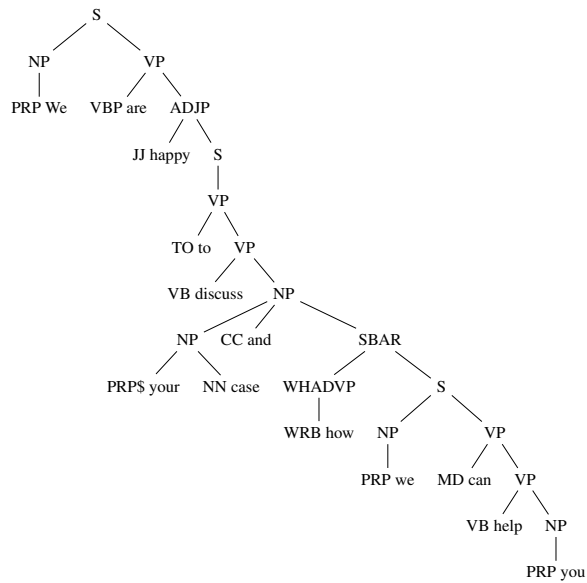

We have provided some examples of parser errors which are discussed in section \ref{sec:eval}. The sentence in Figure~\ref{fig:false-pos} is usually interpreted as a conditional sentence, however the parser represented it as an embedded alternative interrogative clause. On the other hand, the sentence in Figure~\ref{fig:false-neg} is an embedded clause, but the parser represented it as a relative clause.  

\begin{figure}
    \begin{adjustbox}{width=\linewidth}\begin{forest} [S [VP [VB Tweet] [NP [PRP Me]] [SBAR [IN if] [S [NP [PRP You]] [VP [VBP Love] [NP [PRP Me]]]]]]] \end{forest}\end{adjustbox}
    \caption{False-positive parse of a sentence with a conditional}
    \label{fig:false-pos}
\end{figure}
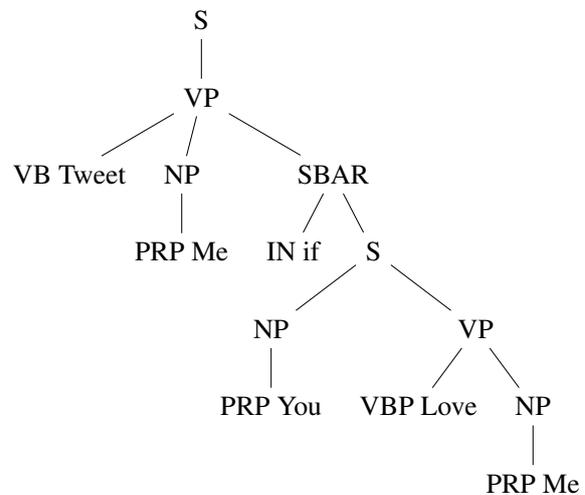

\section{Neg-Raising Generalisation}

Another linguistic generalisation that concerns embedded clauses is Neg-Raising. When syntactically negated, Neg-Raising predicates can take semantically higher
scope than negation \cite{horn1978remarks, gajewski2007neg, ZeijlstraHedde2018DNIN}. Consider (\ref{neg-raise-pred}), which can have the reading as in (\ref{neg-raise-pred}) but can also have the interpretation in (\ref{neg-raise-ec}). Accordingly, \textit{believe} is described as a Neg-Raising predicate. 

\pex 
\a I don't believe John is nice\label{neg-raise-pred}
\a I believe that John isn't nice \label{neg-raise-ec}
\xe

However, many other embedding predicates do not allow this reading. Consider \textit{predict} in (\ref{non-neg-pred}), which does not have the reading in (\ref{non-neg-ec}), so that \textit{predict} is a non-Neg-Raising predicate

\pex
\a I don't predict it will rain \label{non-neg-pred}
\a I predict it will not rain \label{non-neg-ec}
\xe

A generalisation of Neg-Raising predicates that has come from the literature is that they do not select for interrogative embedded clauses \cite{zuber1982semantic, MayrClemens2019Taie, Theiler2019}. Thus, we can see in our dataset if there are genuine counter-examples to this generalisation, by searching for the Neg-Raising predicates that have interrogative embedded clauses. We did this with the predicate \textit{believe} which occurred $497,684$ amount of times in the dataset, of which $19,787$ were interrogative embedded clauses. Looking at a subset of these interrogative embedded clauses, the majority of the instances were not true embedded clauses, which is to be expected as we need to propagate the error of the tool into our analysis. However, there were a few potential true counter-examples like (\ref{counterexample}), which two of the three native speakers amongst the authors found grammatical. However, we leave their analysis for future work. 

\pex\label{counterexample}
\a As we are not omniscient, we can't validate/believe (absolutely) whether God's existence is true/absolute.
\a Ernst was also asked if she believed whether or not the Russia investigation was warranted.
\xe

\end{document}